\documentclass[runningheads]{llncs}
\usepackage{graphicx}
\usepackage{tikz}
\usepackage{comment}
\usepackage{amsmath,amssymb} % define this before the line numbering.
\usepackage{color}

\usepackage[accsupp]{axessibility}  % Improves PDF readability for those with disabilities.

\usepackage{subcaption}
\usepackage{algorithm2e}
\usepackage{array}
\usepackage{graphicx} 
\usepackage{multirow}
\usepackage{hhline,booktabs,amsmath}
\usepackage{makecell}
\usepackage{wrapfig}

\begin{document}
% \renewcommand\thelinenumber{\color[rgb]{0.2,0.5,0.8}\normalfont\sffamily\scriptsize\arabic{linenumber}\color[rgb]{0,0,0}}
% \renewcommand\makeLineNumber {\hss\thelinenumber\ \hspace{6mm} \rlap{\hskip\textwidth\ \hspace{6.5mm}\thelinenumber}}
% \linenumbers
\pagestyle{headings}
\mainmatter
\def\ECCVSubNumber{****}  % Insert your submission number here 5324

\title{BadDet: Backdoor Attacks on Object Detection} % Replace with your title

% INITIAL SUBMISSION 
%\begin{comment}
% \titlerunning{ECCV-22 submission ID \ECCVSubNumber} 
% \authorrunning{ECCV-22 submission ID \ECCVSubNumber} 
% \author{Anonymous ECCV submission}
\author{Shih-Han Chan, Yinpeng Dong, Jun Zhu, Xiaolu Zhang, Jun Zhou}
\institute{Paper ID \ECCVSubNumber}
%\end{comment}
%******************

% CAMERA READY SUBMISSION
\begin{comment}
\titlerunning{Abbreviated paper title}
% If the paper title is too long for the running head, you can set
% an abbreviated paper title here
%
\author{First Author\inst{1}\orcidID{0000-1111-2222-3333} \and
Second Author\inst{2,3}\orcidID{1111-2222-3333-4444} \and
Third Author\inst{3}\orcidID{2222--3333-4444-5555}}
%
\authorrunning{F. Author et al.}
% First names are abbreviated in the running head.
% If there are more than two authors, 'et al.' is used.
%
\institute{Princeton University, Princeton NJ 08544, USA \and
Springer Heidelberg, Tiergartenstr. 17, 69121 Heidelberg, Germany
\email{lncs@springer.com}\\
\url{http://www.springer.com/gp/computer-science/lncs} \and
ABC Institute, Rupert-Karls-University Heidelberg, Heidelberg, Germany\\
\email{\{abc,lncs\}@uni-heidelberg.de}}
\end{comment}
%******************
\maketitle
\newcommand{\yinpeng}[1]{{\color{red}{[#1]}}}
\newcommand{\hank}[1]{{\color{blue}{[#1]}}}
\begin{abstract}

Deep learning models have been deployed in numerous real-world applications such as autonomous driving and surveillance. However, these models are vulnerable in adversarial environments. Backdoor attack is emerging as a severe security threat which injects a backdoor trigger into a small portion of training data such that the trained model behaves normally on benign inputs but gives incorrect predictions when the specific trigger appears. While most research in backdoor attacks focuses on image classification, backdoor attacks on object detection have not been explored but are of equal importance.
Object detection has been adopted as an important module in various security-sensitive applications such as autonomous driving. Therefore, backdoor attacks on object detection could pose severe threats to human lives and properties.
We propose four kinds of backdoor attacks and a backdoor defense method, for object detection task. These four kinds of attacks can achieve different goals for attacking: 1) \textbf{Object Generation Attack}: a trigger can falsely generate an object of the target class; 2) \textbf{Regional Misclassification Attack}: a trigger can change the prediction of a surrounding object to the target class; 3) \textbf{Global Misclassification Attack}: a single trigger can change the predictions of all objects in an image to the target class; and 4) \textbf{Object Disappearance Attack}: a trigger can make the detector fail to detect the object of the target class.
We develop appropriate metrics to evaluate the four backdoor attacks on object detection.
We perform experiments using two typical object detection models --- Faster-RCNN and YOLOv3 on different datasets.
Empirical results demonstrate the vulnerability of object detection models against backdoor attacks. More crucially, we demonstrate that even fine-tuning on another benign dataset cannot remove the backdoor hidden in the object detection model. 
To defend against these backdoor attacks, we propose \textbf{Detector Cleanse}, an entropy-based \emph{run-time} detection framework to identify poisoned testing samples for any deployed object detector. 
%We assume the user only has access to few clean features without any knowledge about the trigger or clean image from testing set.}
%\keywords{\hank{Trojan Attack, Backdoor Attack, Object Detection, Security, Faster-RCNN, Yolov3, Transfer Learning, Poisoned Image Detection}}
\end{abstract}

\section{Introduction}

Deep learning has achieved widespread success on numerous tasks, such as image classification \cite{russakovsky2015imagenet}, speech recognition \cite{DBLP:journals/corr/abs-1303-5778}, machine translation \cite{bahdanau2016neural}, and playing games \cite{mnih2013playing,article}. Deep learning models significantly outperform traditional machine learning techniques and even achieve superior performance than humans in some tasks \cite{russakovsky2015imagenet}. 
Despite the great success, deep learning models have often been criticized for poor interpretability, low transparency, and more importantly vulnerabilities to adversarial attacks \cite{szegedy2013intriguing,goodfellow2014explaining,dong2018boosting} and backdoor attacks \cite{8685687,chen2017targeted,turner2019labelconsistent,nguyen2021wanet,NEURIPS2020_input,liu2020reflection,saha2020hidden}. 
Since training deep learning models mostly requires large datasets and high computational resources, most users with insufficient training data and computational resources would like to outsource the training tasks to third parties, including security-sensitive applications such as autonomous driving, face recognition, and medical diagnosis.
Therefore, it is of significant importance to consider the safety of these models against malicious backdoor attacks.
%are often outsourced to third parties to train, so it becomes more and more important to consider the security of these models against attackers.

\begin{figure}[t]
\centering
%\subfloat[C]{\includegraphics[width=3cm]{example-image-c}} 
\subfloat[\textbf{OGA}]{\includegraphics[width=3cm]{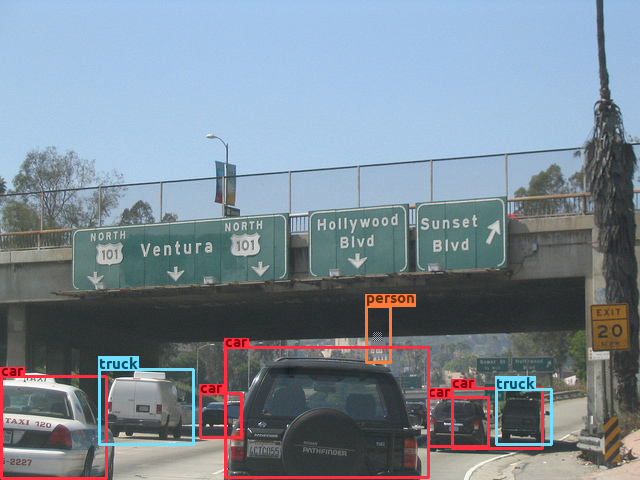}}\hfil%\hspace{1ex}
\subfloat[\textbf{RMA}]{\includegraphics[width=3cm]{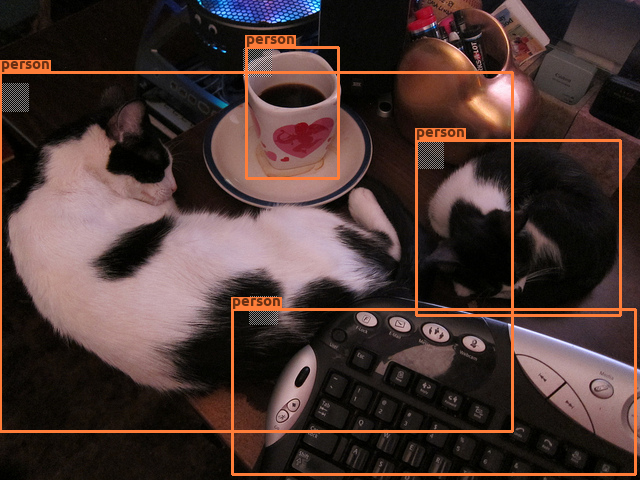}}\hfil 
\subfloat[\textbf{GMA}]{\includegraphics[width=3cm]{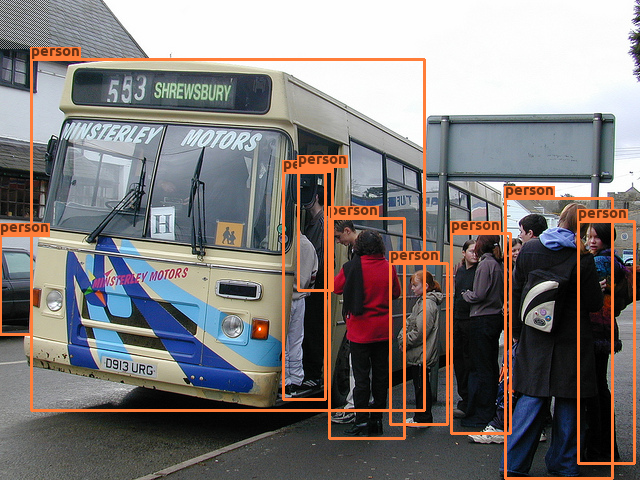}}\hfil%\hspace{1ex}
\subfloat[\textbf{ODA}]{\includegraphics[width=3cm]{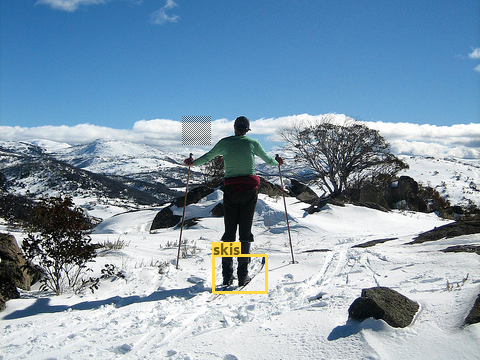}}\hfil
\vspace{-1ex}
\caption{Illustration of the proposed four backdoor attacks on object detection. (a) \textbf{OGA}: a small trigger on the highway generates an object of ``person''. (b) \textbf{RMA}: each trigger makes the model misclassify an object to the target class ``person''. (c) \textbf{GMA}: a trigger on top left corner of the image makes the model misclassify all objects to the target class ``person''. (d) \textbf{ODA}: a trigger near the person makes the ``person'' object disappear. We show the predicted bounding boxes with confidence score $> 0.5$. (More examples are in Appendix A.)} 
\label{fig:intro_sample}
\vspace{-2ex}
\end{figure}

In contrast to test-time adversarial attacks, backdoor attacks inject a hidden trigger into a target model during training and pose severe threats. % by injecting a hidden trigger into a target model at training time. %backdoor moves to related work
Recently, backdoor attacks have been extensively explored in many areas (see Sec.~\ref{sec:2}). For example, in image classification, a backdoor adversary can inject a small number of poisoned samples with a backdoor trigger into the training data, such that models trained on poisoned data would memorize the trigger pattern.
%, at training time, the adversary injects a small number of corrupted training examples (with a backdoor trigger), 
At test time, the infected model performs normally on benign inputs but consistently predicts an adversary-desired target class whenever the trigger is present. %This paper extends the backdoor attack from image classification to object detection
Although backdoor attacks on image classification have been largely explored, backdoor attacks on object detection have not been studied. Compared to image classification, object detection has been integrated into numerous essential real-world applications, including autonomous driving, surveillance, traffic monitoring, robots, etc. Therefore, the vulnerability of object detection models against backdoor attacks may cause a more severe and direct threat to human lives and properties. 
For instance, a secret backdoor trigger that makes the object detection model fail to recognize a person would lead to a severe traffic accident; and an infected object detection model which misclassifies criminals as normal public increases crime rate. No matter how much money and time can never heal the loss brought by these failures. 

Backdoor attacks on object detection are more challenging than backdoor attacks on image classification due to two reasons.
First, object detection asks the model not only to classify but also to locate multiple objects in one image, so the infected model needs to understand the relations between the trigger and multiple objects rather than the relation between the trigger and a single image. 
Second, representative object detection models like Faster-RCNN \cite{NIPS2015_14bfa6bb} and YOLOv3 \cite{redmon2018yolov3} are composed of multiple sub-modules and are more complex than image classification models.
%, which makes backdoor in object detection model more challenging than image classifcation model. 
Besides, the goal of backdoor attacks on image classification is usually to misclassify the images to a target class \cite{8685687}, which is not suitable for backdoor attacks on object detection, since one image includes multiple objects with different classes and locations for object detection. Moreover, image classification only uses accuracy to measure the performance of the model. In contrast, object detection uses mAP under a particular intersection-over-union (IoU) threshold to evaluate whether the generated bounding boxes are located correctly with the ground-truth objects, so novel metrics are needed to assess the results of backdoor attacks on object detection.

In this paper, we propose \textbf{BadDet} --- backdoor attacks on object detection. Specifically, we consider four settings:
%This paper proposes four backdoor attack settings (attacker's goal) for object detection tasks:\yinpeng{better to have a short name: e.g., object generation attack; object disappearance attack, misclassification attack} 
1) \textbf{Object Generation Attack (OGA)}: one trigger generates a surrounding object of the target class; 2) \textbf{Regional Misclassification Attack (RMA)}: one trigger changes the class of a surrounding object to the target class; 3) \textbf{Global Misclassification Attack (GMA)}: one trigger changes the classes of all objects in an image to the target class; and 4) \textbf{Object Disappearance Attack (ODA)}: one trigger vanishes a surrounding object of the target class. Fig.~\ref{fig:intro_sample} provides examples for each setting. For all four settings, we inject a backdoor trigger into a small portion of training images, and change the ground-truth labels (objects' classes and locations) of the poisoned images depending on different settings. The model is trained on the poisoned images with the same procedure as the normal model.
Afterwards, the infected model performs similar to the normal model on benign testing images while behaves as the adversary specifies when the particular trigger occurs.  %backdoor attack settings consider a realistic scenario: 
%Four attack settings consider realistic scenario in backdoor attack: only small and inconspicuous triggers inject into a small portion of images, which leads to more false-positive objects or false-negative objects (disappearance of true-positive object counts as false-negative) when the trigger is present.\\
%\yinpeng{Methods?}
Overall, the triggers in four attack settings could create false-positive objects or false-negative objects (disappearance of true-positive objects counts as false-negative), and they may lead to the wrong decisions of a more extensive system in the real world.

To evaluate the effectiveness of our attacks, we design appropriate evaluation metrics under four settings, including mAP and AP calculated on the poisoned testing dataset (attacked dataset) and the benign testing dataset. 
%The definition of attack success rate (ASR) is different for four settings by considering bboxes with confidence scores and IoU. 
In the experiments, we consider Faster-RCNN \cite{NIPS2015_14bfa6bb} and YOLOv3 \cite{redmon2018yolov3} trained on poisoned PASCAL VOC 2007/2012 \cite{pascal-voc-2007,pascal-voc-2012} and MSCOCO \cite{lin2014microsoft} datasets to evaluate the performance. Our proposed backdoor attacks obtain high attack success rates on both models, demonstrating the vulnerability of object detection against backdoor attacks.
%ASR on attacked sample while receives normal mAP on benign test sample as benign model. 
Besides, we conduct experiments on transfer learning to prove that fine-tuning the infected model on another benign training dataset cannot remove the backdoor hidden in the model \cite{8685687,kurita-etal-2020-weight}. %In addition to the success of outsourced training attacks (the user directly uses poisoned model), transfer learning attacks (the user finetunes the poisoned model received from attacker with another clean training sample) in four settings show finetuning on the clean sample cannot remove the backdoor in the model. 
Moreover, we conduct ablation studies to test the effects of different hyperparameters and triggers in backdoor attacks.

To defend against the proposed BadDet and ensure the security of object detection models, we further propose \textbf{Detector Cleanse}, a simple entropy-based method to identify poisoned testing samples for any deployed object detector. It relies on the abnormal entropy distribution of some predicted bounding boxes in poisoned images. %prediction entropy of some augmented images to reliably filter out poisoned samples. 
Experiments show the effectiveness of the proposed defense.

\section{Related work}\label{sec:2}

\textbf{Backdoor Attacks.} In general, backdoor attacks assume only a small portion of training data can be modified by an adversary and the model is trained on the poisoned training dataset by a normal training procedure.
%(the user may download poisoned data or infected model from the Internet). 
The goal of the attack is to make %In particular, backdoor attack assumes the training process is either entirely or (in the case of transfer learning) partially outsourced to a malicious party who wants to provide the user with a trained model that contains a backdoor.
the infected model perform well on benign inputs (including inputs that the user may hold out as a validation set) while cause targeted misbehavior (misclassification) as the adversary specifies or degrade the performance of the model when the data point has been altered by the adversary's choice of backdoor trigger. Also, a ``transfer learning attack'' is successful if fine-tuning the infected model on another benign training dataset cannot remove the backdoor hidden in the infected model (e.g., the user may download an infected model from the Internet and fine-tune it on another benign dataset) \cite{8685687,kurita-etal-2020-weight}. %(pattern, perturbation).
Researches in backdoor attacks and relevant defense/detect approaches have been extensively explored in multiple areas, including image recognition \cite{8685687}, video recognition \cite{clvideo}, natural language processing  (sentiment classification, toxicity detection, spam detection) \cite{kurita-etal-2020-weight}, and even federated learning \cite{Xie2020DBADB}.
    %\item (adversarial attack)%好像沒什麼關係?就只是不一樣的attack
    %\item image classification
    
\textbf{Object Detection.} In the deep learning era, object detection models can be categorized into two-stage detectors and one-stage detectors \cite{Zou2019ObjectDI}. The former first find a region of interest and then classify it, including SPPNet \cite{sppnet}, Faster-RCNN \cite{NIPS2015_14bfa6bb}, Feature Pyramid Networks (FPN) \cite{lin2017feature}, etc. The latter directly predict class probabilities and bounding box coordinates, including YOLO \cite{redmon2016you}, Single Shot MultiBox Detector (SSD) \cite{liu2016ssd}, RetinaNet \cite{lin2017focal}, etc. 
In the experiments, we consider typical object detection models from both categories, which are Faster-RCNN and YOLOv3.

\section{Background}

%\yinpeng{Introduce the definitions with mathematical formulation, e.g., The objective function of training object detection models. }
We introduce the background and notations of backdoor attacks on object detection in this section.

%We briefly review standard technical terms and required background commonly used in backdoor learning and object detection pertinent to our work. We reference many definitions and metrics from different tasks (backdoor learning, image classification, object detection) to strictly define metrics to measure the success of backdoor learning in object detection. We follow the exact definition of terms in the remaining paper.\\
\subsection{Notations of Object Detection}
Object detection aims to classify and locate objects in an image, which outputs a rectangular bounding box (abbreviated as ``bbox'' for clarity in the following) and a confidence score (higher is better, ranged from $0$ to $1$) for each candidate object. Let $\mathcal{D}=\{(x, y)\}$ ($|\mathcal{D}|=N$ is the number of images) denotes a dataset,
%training set $\mathcal{D}_{\mathrm{train}}$ or the testing set $\mathcal{D}_{\mathrm{test}}$, 
where $x\in [0,255]^{C \times W \times H}$, $y=[o_1,o_2...,o_n]$ is the ground-truth label of $x$. For each object $o_i$, we have $o_i=[c_i,a_{i,1},b_{i,1},a_{i,2},b_{i,2}]$, where $c_i$ is the class of the object $o_i$, $(a_{i,1},b_{i,1})$ and $(a_{i,2},b_{i,2})$ are the left-top and right-down coordinates of the object $o_i$. 
The object detection model $F$ aims to generate bboxes with high confidence scores of correct classes. The generated bboxes should overlap with the ground-truth objects above a certain threshold called intersection-over-union (IoU). Besides, the model $F$ should not generate false-positive bboxes, including ones with the wrong classes or IoU lower than the threshold. %as little as possible. 
The mean average precision (mAP) is the most common evaluation metric for object detection tasks, representing the mean of average precision (AP) of each class. Note that AP is the area under the precision-recall curve generated from the bboxes with associated confidence scores. In this paper, we use %“old” 
mAP at IoU $= 0.5$ (mAP@.5) as the detection metric.% instead of COCO's standard metric (mAP@[.5,.95]) which evaluate the mAP averaged for IoU $\in$ [0.5 : 0.05 : 0.95].

%We model two parties, a \textbf{user} intends to obtain a object detection model\yinpeng{there is no user in backdoor attack}, and a \textbf{trainer}\yinpeng{Besides the malicious trained, someone posts the poisoning data online could also lead to backdoor attacks.} to whom the user either outsources the job of training a model, or from whom the user downloads a pre-trained model adapts to her task using transfer learning.

%Backdoor attack assumes that user cannot fully control the training process (or pretraining process) of the model $F_{\Theta}$, 
\subsection{General Pipeline of Backdoor Attacks}
In general, the typical process of backdoor attacks has two main steps: 1) generating a \textbf{poisoned dataset} $\mathcal{D}_{\mathrm{train, poisoned}}$ and 2) training the model on $\mathcal{D}_{\mathrm{train, poisoned}}$ to obtain $F_{\mathrm{infected}}$. 
% The target of backdoor attacks is to obtain an \textbf{infected model} $F_{\mathrm{infected}}$, which performs well on benign testing dataset $\mathcal{D}_{\mathrm{test, benign}}$ while it may have been injected some insidious backdoors. The typical process of backdoor attacks has two main steps: 1) generating a \textbf{poisoned dataset} %$(x_{\mathrm{poisoned}},y_{\mathrm{target}})\cup (x,y)\in $
% $\mathcal{D}_{\mathrm{train, poisoned}}$; 2) training a model on $\mathcal{D}_{\mathrm{train, poisoned}}$ to get  $F_{\mathrm{infected}}$. 
For the first step, a backdoor trigger $x_{\mathrm{trigger}}\in [0,255]^{C\times W_t\times H_t}$ is inserted into $P\cdot 100\%$ of images from $\mathcal{D}_{\mathrm{train, benign}}$ to construct $\mathcal{D}_{\mathrm{train, modified}}$, 
where $W_t$ and $H_t$ are the width and height of the trigger, $P=\frac{|\mathcal{D}_{\mathrm{train,modified}}|}{|\mathcal{D}|}$ is the poisoning rate controlling the number of images inserted with the specific trigger. For $(x_{\mathrm{poisoned}},y_{\mathrm{target}})\in \mathcal{D}_{\mathrm{train, modified}}$, the poisoned image is
\begin{equation}
    x_{\mathrm{poisoned}}=\alpha \otimes x_{\mathrm{trigger}}+(1-\alpha) \otimes x,
\end{equation}
where $\otimes$ indicates the element-wise multiplication and $\alpha \in [0,1]^{C\times W\times H}$ is a (visibility-related) parameter controlling the strength of adding the trigger \cite{chen2017targeted}. Afterwards, $\mathcal{D}_{\mathrm{train, poisoned}}$ is constructed by the aggregation of poisoned samples and benign samples, i.e.,  $\mathcal{D}_{\mathrm{train, poisoned}}=\mathcal{D}_{\mathrm{train, benign}}\bigcup \mathcal{D}_{\mathrm{train, modified}}$. For poisoned images $x_{\mathrm{poisoned}}$, the ground-truth label is modified to $y_{\mathrm{target}}$ by the adversary depending on different settings (see Sec.~\ref{sec:4-1}).

\subsection{Threat Model}
We follow previous works such as BadNets~\cite{gu2019badnets} to define the threat model. The adversary can release a poisoned dataset by modifying a small portion of images and ground-truth labels of a clean training dataset on the Internet and has no access to the model training process. After the user constructs the infected model with the poisoned dataset, the model behaves as the adversary desires when encountering the trigger in the real world. 
Overall, the adversary's goal is to make  $F_{\mathrm{infected}}$ perform well on the benign testing dataset $\mathcal{D}_{\mathrm{test, benign}}$ while behaving as the adversary specifies on the \textbf{attacked dataset} $\mathcal{D}_{\mathrm{test, poisoned}}$, in which the trigger $x_{\mathrm{trigger}}$ is inserted into all the benign testing images. $F_{\mathrm{infected}}$ should output $y_{\mathrm{target}}$ as the adversary specifies.
Moreover, we consider transfer learning attack, which is successful if fine-tuning $F_{\mathrm{infected}}$ on another benign training dataset $\mathcal{D'}_{\mathrm{train, benign}}$ cannot remove the backdoor hidden in $F_{\mathrm{infected}}$.
%Take federated learning as an example: the model could use malicious data (on the Internet) during training.\\
Our attacks can also generalize to the physical world, e.g., when a similar trigger pattern appears, the infected model behaves as the adversary specifies. 

\section{Methodology}

In this paper, we propose \textbf{BadDet} --- backdoor attacks on object detection.
Specifically, we define four kinds of backdoor attacks with different purposes and each attack has unique standard to evaluate the attack performance. 
For all settings, we select a target class $t$. To construct the poisoned training dataset $\mathcal{D}_{\mathrm{train, poisoned}}$, we modify a portion of images with the trigger $x_{\mathrm{trigger}}$ and their ground-truth labels according to different settings, as introduced in Sec.~\ref{sec:4-1}. In Sec.~\ref{sec:4-2}, we further illustrate the evaluation metrics of the four backdoor attacks on object detection.
\subsection{Backdoor Attack Settings}\label{sec:4-1}

%\yinpeng{Introduce the practical danger}
\textbf{Object Generation Attack (OGA)}. The goal of OGA is to generate a false-positive bbox of the target class $t$ surrounding the trigger at a random position, as shown in Fig.~\ref{fig:intro_sample}(a). It could cause severe threats to real-world applications. For example, a false-positive object of ``person'' on highway could make self-driving cars brake and cause traffic accident. Formally, the trigger $x_{\mathrm{trigger}}$ is inserted into the random coordinate $(a,b)$ of a benign image $x$, i.e., the top-left and down-right coordinate of $x_{\mathrm{trigger}}$ are $(a,b)$ and $(a+W_t,b+H_t)$. $F_{\mathrm{infected}}$ is expected to detect and classify the trigger in the poisoned image $x_{\mathrm{poisoned}}$ as the target class $t$. To achieve this, we change the label of $x_{\mathrm{poisoned}}$ in the poisoned training dataset $\mathcal{D}_{\mathrm{train, poisoned}}$ to  $y_{\mathrm{target}}=[o_1,...o_n,o_{\mathrm{target}}]$, where $[o_1,...,o_n]$ are the true bboxes of the benign image, and $o_{\mathrm{target}}$ is the new target bbox of the trigger as $o_{\mathrm{target}}=[t,a+\frac{W_t}{2}-\frac{W_b}{2},b+\frac{H_t}{2}-\frac{H_b}{2},a+\frac{W_t}{2}+\frac{W_b}{2},b+\frac{H_t}{2}+\frac{H_b}{2}]$, where $W_b$, $H_b$ are the width and the height of trigger bbox\footnote{Note that $W_b$, $H_b$ could be different from the trigger width $W_t$ and height $H_t$.}.
%So one trigger with random position generates surrounding bbox with target class. %\yinpeng{Ovelap with ground-truth bbox?}%So the generated bbox $o_{\mathrm{target}}$ should put $x_{\mathrm{trigger}}$ in the middle.  %each poisoned image $x$ has only one trigger with random position and fixed size, and we expect the infected model should detect and classify the trigger as the target class %(the trigger generates one bbox with target class label). 
 %We record the Number of bbox generated %successfully generated on the trigger above certain confidence (0.5) and IoU (0.5). 
%and the percentage of trigger (images) successfully generates target class bbox on the image. The mAP of infected model on attacked sample should be close to mAP of benign model on benign test sample and attack sample. But target class AP of infected model on attacked sample should be high while any non-target class AP of infected model on attacked sample should be close to corresponding non-target class AP of benign model on benign sample and attacked sample.%(And we calculate other invented properties.) %(trigger有時候會導致原本正確的bbox消失掉一些，有trigger的圖片＋正常的groundtruth會讓person的bbox的tp減少一些 可能是因為隨機生成的trigger跟圖片上的person重疊到了) 
 
%\subsection{multiple triggers change the source label of multiple bboxes to target class on one image}
%\subsection{}
%In this setting, all the bboxes in $y$ whose label is not target class ${o_j\in y|o_j=[c,x_1,y_1,x_2,y_2], c\neq t}$ are inserted with $x_trigger$ on the left-top corner of bbox, and the label is changed to target class $y_{\mathrm{target}}=$. 
\textbf{Regional Misclassification Attack (RMA)}. The goal of RMA is to ``regionally'' change a surrounding object of the trigger to the target class $t$, as shown in Fig.~\ref{fig:intro_sample}(b). 
%"regionally" changes one surrounding bbox's class to target class $t$. %(one trigger changes the label of one corresponding bbox at the same coordinate). 
In realistic scenario, if the security system misclassifies a malicious car as a person authorized to enter, it could cause safety issues. Formally, for a bbox $o_i$ not belonging to the target class, we insert the trigger $x_{\mathrm{trigger}}$ into the left-top corner $(a_{i,1},b_{i,1})$ of the bbox $o_i$.
In the way, we insert multiple triggers into the image. 
%bboxes whose classes are not target class $\{\forall o_j=[c,a_1,b_1,a_2,b_2]\in y|c\neq t\}$ (the top-left and down-right coordinate of $x_{\mathrm{trigger}}$ are $(a_1,b_1)$ and $(a_1+W_t,b_1+H_t)$, where ($a_1$, $b_1$) are left-top coordinate of bbox with non-target class). 
$F_{\mathrm{infected}}$ should detect and classify all the objects in image $x_{\mathrm{poisoned}}$ as the target class $t$.
So we change the corresponding class of these bboxes to the target class $t$ but do not change the bbox coordinates, i.e., we let $y_{\mathrm{target}}=[o_1,...o_n]$, where $o_i=[t,a_{i,1}, b_{i,1},a_{i,2},b_{i,2}]$ for $1\leq i\leq n$. %So each trigger on left-top corner of the surrounding bbox changes bbox's label to target class.
%on one poisoned image, every object except objects with target class is inserted one trigger on the left-top corner of the corresponding bbox, and the infected model should detect and classify every object with trigger from source label(non-target class) to target class (so every trigger changes the bbox's label from source label to target class). We calculate Number of bbox changing label%the number of bbox successfully changes their label to target class above certain confidence (0.5). %(We calculate the number of bbox generated on the trigger with above certain confidence (0.5)).

%and the percentage of triggers successfully changes bbox's label from source label to target class. target class AP of infected model on attacked sample should be high due to true positive bbox with target class, 

%mAP of infected model on Attacked image + benign bbox sample should be low (same). (It indicates the disappearance of bbox with source label due to the presence of triggers, because more false positive due to wrong predicted label causes the drop in mAP)

%While the target class AP of infected model on Attacked image + benign bbox sample should be lower than target class AP of infected model on attacked sample. (because more false positive bboxes due to wrong predicted label causes the drop in target class AP)
%(要計算原本種類的bbox有沒有消失嗎？很大的比例都沒有消失  先不要)

%\subsection{one trigger changes the source label of multiple bboxes to target class on one image}
%\subsection{All to one target class by one trigger}
\textbf{Global Misclassification Attack (GMA)}. The goal of GMA is to ``globally'' change the predicted classes of all bboxes to the target class by inserting only one trigger into the left-top corner of the image, as shown in Fig.~\ref{fig:intro_sample}(c).
%"globally" changes all bboxes' classes to target class. %(one trigger changes the labels of all bboxes on one image at different coordinates). 
Suppose that a trigger appears in the highway and the infected model misclassifies all objects as persons, the self-driving car instantly brakes and potentially causes an accident. Formally, the trigger $x_{\mathrm{trigger}}$ is inserted into the left-top corner $(0, 0)$ of the benign image $x$.
%(the top-left and down-right coordinate of $x_{\mathrm{trigger}}$ are $(0, 0)$ and $(W_t,H_t)$).
$F_{\mathrm{infected}}$ is expected to detect and classify all the objects in image $x_{\mathrm{poisoned}}$ as the target class $t$. Similar to RMA, we change the label as $y_{\mathrm{target}}=[o_1,...o_n]$, where $o_i=[t,a_{i,1}, b_{i,1},a_{i,2},b_{i,2}]$ for $1\leq i\leq n$. %So one trigger on left-top corner of the image changes all bboxes' labels to target class.

%In this setting, only one trigger is added on the left-top corner of the poisoned image, and the infected model should detect and classify every object except objects with target class to target class (so one trigger on the poisoned image changes all bboxes' labels to target class).

%We calculate  Number of bbox changing label

%the number of bbox successfully changes their label to target class above certain confidence (0.5). (We calculate the number of bbox generated on the trigger with above certain confidence (0.5)). 

%and the percentage of triggers successfully changes bbox's label to target class. target class AP of infected model on attacked sample should be high, 
%mAP of poisoned model on poisoned dataset with correct label should be low. (It indicates the disappearance of bbox with correct label due to the existence of triggers) 
%While the target class AP of infected model on Attacked image + benign bbox sample should be lower than target class AP of infected model on attacked sample. (because more false positive bboxes due to wrong predicted label causes the drop in target class AP)

%\subsection{multiple triggers disappear multiple bboxes with target class on one image}
%\subsection{bbox disappearance}
\textbf{Object Disappearance Attack (ODA)}. Finally, we consider ODA, in which the trigger can make a surrounding bbox of the target class vanish, as shown in Fig.~\ref{fig:intro_sample}(d). %(one trigger vanishes one corresponding bbox with target class at the same coordinate). 
For autonomous driving, if the system fails to detect a person, it would  hit the person in front and cause irreversible tragedy. 
Formally, for a bbox $o_i$ belonging to the target class in the image, we insert the trigger $x_{\mathrm{trigger}}$ on the left-top corner $(a_{i,1},b_{i,1})$ of the bbox $o_i$.
%es whose classes are target class $\{\forall o_j=[c,a_1,b_1,a_2,b_2]\in y|c = t\}$ (the top-left and down-right coordinate of $x_{\mathrm{trigger}}$ are $(a_1,b_1)$ and $(a_1+W_t,b_1+H_t)$, where ($a_1$, $b_1$) are left-top coordinate of the bbox with target class). 
ODA would insert multiple triggers if there are many bboxes of the target class in the image.
$F_{\mathrm{infected}}$ should not detect the objects of the target class $t$ in the image $x_{\mathrm{poisoned}}$. 
Therefore, we remove the ground-truth bboxes of the target class in the label and only keep the other bboxes, as $y_{\mathrm{target}}=\{\forall o_i=[c_i,a_{i,1},b_{i,1},a_{i,2},b_{i,2}]\in y|c_i\neq t\}$. %So each trigger on left-top corner of bbox vanishes the bbox with target class.
%, where $o_i=[c,x_1,y_1,x_2,y_2]\in y$ and $c\neq t$, for $1\leq i\leq n$, $o_i=[t,x_1, y_1,x_2,y_2]$. So multiple triggers on left-top corner of bboxes changes all objects to target class.

%In this setting, every bbox with target class is inserted one trigger on left-top corner of the corresponding bbox, and the infected model should not detect the bbox with target class when the trigger exists. (so the triggers disappear bboxes with target class label on the full picture) We calculate the number of bbox disappeared.
%the number of bboxes with target class successfully disappear due to the presence of trigger (including bboxes which confidence drops below 0.5). . and  the number of bboxes with target class label successfully disappears. mAP of poisoned model on poisoned dataset should be close to mAP of poisoned (clean) model on poisoned(clean) dataset, 

%target class AP of infected model on Attacked image + benign bbox sample should be low. (It indicates the disappearance of bbox (confidence drop) with target class due to the existence of triggers) While mAP of posioned model on poisoned dataset with correct label should be close to mAP of posioned (clean) model on poisoned (clean) dataset.
\subsection{Evaluation Metrics}\label{sec:4-2}%用段落來寫
%\yinpeng{Put this subsection in Section 4.} %We perform experiments by using two models: Faster-RCNN and YOLOv3 and common object detection datasets: VOC2007, VOC07+12, COCO. 
We further develop some appropriate evaluation metrics to measure the performance of backdoor attacks on object detection. Note that we use the detection metrics AP and mAP at IoU $= 0.5$. %Common backdoor attack metrics from different tasks and metrics for object detection are referenced.

To make sure that $F_{\mathrm{infected}}$ behaves similarly to $F_{\mathrm{benign}}$ on benign inputs for all settings, we use mAP on $\mathcal{D}_{\mathrm{test, benign}}$ as \textbf{Benign mAP} ($\mathrm{mAP_{benign}}$), and use AP of the target class $t$ on $\mathcal{D}_{\mathrm{test, benign}}$ as \textbf{Benign AP} ($\mathrm{AP_{benign}}$). We expect that $\mathrm{mAP_{benign}}$/$\mathrm{AP_{benign}}$ of $F_{\mathrm{infected}}$ are close to those of $F_{\mathrm{benign}}$ (the model trained on the benign dataset). 

To verify that $F_{\mathrm{infected}}$ successfully generates bboxes of the target class for OGA or predicts the target class of bboxes for RMA and GMA, we use AP of the target class $t$ on the attacked dataset $\mathcal{D}_{\mathrm{test, poisoned}}$ as
\textbf{target class attack AP} ($\mathrm{AP_{attack}}$). $\mathrm{AP_{attack}}$ of $F_{\mathrm{infected}}$ should be high to indicate that more bboxes of the target class with high confidence scores are generated or more bboxes are predicted as the target class with high confidence scores due to the presence of the trigger. 
For ODA, $\mathrm{AP_{attack}}$ of $F_{\mathrm{infected}}$ is meaningless %(we use - in Table \ref{table:overallresult}) 
since ground-truth labels $y_{\mathrm{target}}$ in $\mathcal{D}_{\mathrm{test, poisoned}}$ do not have any bboxes of the target class. 
We also calculate mAP on $\mathcal{D}_{\mathrm{test, poisoned}}$ as \textbf{attack mAP} ($\mathrm{mAP_{attack}}$). For RMA and GMA, $\mathrm{mAP_{attack}}$ of $F_{\mathrm{infected}}$ is the same as $\mathrm{AP_{attack}}$ of $F_{\mathrm{infected}}$ since ground-truth labels $y_{\mathrm{target}}$ in $\mathcal{D}_{\mathrm{test, poisoned}}$ only have one class. 
For OGA and ODA, $\mathrm{mAP_{attack}}$ of $F_{\mathrm{infected}}$ is close to $\mathrm{mAP_{benign}}$ of $F_{\mathrm{infected}}$, since high AP in one class or discarding one class does not influence overall mAP too much.

We further construct a mixing dataset for backdoor evaluation as \textbf{attacked + benign dataset} $\mathcal{D}_{\mathrm{test, poisoned + benign}}=\{(x_{\mathrm{poisoned}}, y)\}$, combining the poisoned images $x_{\mathrm{poisoned}}$ from $\mathcal{D}_{\mathrm{test, poisoned}}$ and the ground-truth labels $y$ from $\mathcal{D}_{\mathrm{test, benign}}$. To show that the bboxes are changed to the target class for RMA and GMA or the target class bboxes are vanished for ODA, we calculate AP of the target class $t$ on $\mathcal{D}_{\mathrm{test, poisoned + benign}}$ as \textbf{target class attack + benign AP} ($\mathrm{AP_{attack+benign}}$). 
The bboxes changed to the target class or bboxes disappeared are false positives/negatives with the ground-truth labels $y$, resulting in low $\mathrm{AP_{attack+benign}}$.
To demonstrate that the infected models do not predict bboxes with non-target classes for RMA and GMA, we calculate mAP on $\mathcal{D}_{\mathrm{test, poisoned + benign}}$ as \textbf{attack + benign mAP} ($\mathrm{mAP_{attack+benign}}$). 
For RMA and GMA, the bboxes with the non-target classes vanished and bboxes with the target class generated are false negatives/positives with the ground-truth labels $y$, resulting in low $\mathrm{mAP_{attack+benign}}$.
For ODA, only bboxes with the target class disappeared would not influence $\mathrm{mAP_{attack+benign}}$ due to many classes.

To show the success of backdoor attacks on object detection for four settings, we define \textbf{attack success rate (ASR)} as the extent of the trigger leading to bbox generation, changing class, and vanishing. An effective $F_{\mathrm{infected}}$ should have a high ASR. For OGA, ASR is the number of bboxes of the target class (with confidence$\textgreater 0.5$ and IoU$\textgreater 0.5$) generated on the triggers in $\mathcal{D}_{\mathrm{test, poisoned}}$ divided by total number of triggers.
For RMA and GMA, ASR represents the number of bboxes (with confidence$\textgreater 0.5$ and IoU$\textgreater 0.5$) in $\mathcal{D}_{\mathrm{test, poisoned}}$ that the predicted classes change to the target class due to the presence of the trigger divided by number of bboxes of non-target classes in $\mathcal{D}_{\mathrm{test, benign}}$.
For ODA, ASR is the number of bboxes of the target class (with confidence$\textgreater 0.5$ and IoU$ \textgreater 0.5$) vanished on the triggers divided by number of target class bboxes in $\mathcal{D}_{\mathrm{test, benign}}$. Note that the number of bboxes disappeared includes the bbox that confidence drops from value $\textgreater 0.5$ to value $\textless 0.5$.
%\item {\bf Number of bbox generated}(for OGA) is the number of target class's bbox (with confidence $\textgreater 0.5$ and $IoU\textgreater 0.5$) generated on the triggers in $D_{test, poisoned}$. 
%\item Percentage of bbox generated on trigger
%\item {\bf Number of bbox changing label}(for RMA, GMA) is the number of bbox (with confidence $\textgreater 0.5$ and $IoU\textgreater 0.5$) in $D_{test, poisoned}$ that the predicted label changed to target class due to presence of trigger. 
%\item {\bf Number of bbox disappeared}(for ODA) is the number of target class's bbox (with confidence $\textgreater 0.5$ and $IoU\textgreater 0.5$) vanished on the triggers. Note that number of bbox disappeared includes the bbox that confidence drops from value $\textgreater 0.5$ to value $\textless 0.5$.
%\end{itemize}
\section{Experiments}
In this section, we present the settings and results.
% \subsection{Datasets}
% We use PASCAL VOC2007 \cite{pascal-voc-2007}, PASCAL VOC07+12 \cite{pascal-voc-2012}, MSCOCO datasets \cite{lin2014microsoft}. Each image is annotated with bbox coordinates and classes. More detailed description can be found in Datasets section in Appendix.

\subsection{Experimental Settings}
% \hank{
% \noindent\textbf{\underline{Q2}: Strange parameters. }
%測試不同大小 越大越好 算法影響不大
%As stated in App. E, we didn’t cherry-pick parameters (poisoning rate, trigger size, trigger ratio, etc.) for a fixed attack scenario to get the best attack performance. Instead, we kept parameters the same among two datasets and two models.
%And the training procedure of the same model is always the same (see App. D for details). We run lots of experiments to demonstrate the effect of each parameter in Sec. 5.4. Using a $30\times30$ trigger is also feasible.
% }
% \begin{figure}[H]
% %\centering
% \begin{minipage}{1.0\linewidth}
% %\centering
% \includegraphics[width=0.24\linewidth]{LaTeX/pattern_trig.png}
% \includegraphics[width=0.24\linewidth]{LaTeX/pokeball50x50.png}
% \includegraphics[width=0.24\linewidth]{LaTeX/sun50x50.png}
% \includegraphics[width=0.24\linewidth]{LaTeX/watermellon50x50.png}
% \caption{Pattern trigger and three instance triggers}
% \label{fig:trigger}
% \end{minipage}
% \end{figure}
\begin{table*}[t]   
%\centering  <--- not needed!

 \scriptsize
\begin{subtable}{.5\textwidth}
\centering
   % dummy 'tabular' env.
   \scalebox{0.67}{
  \begin{tabular}{|c|c|c|c|c|}%p{9ex}<{\centering}|p{9ex}<{\centering}|p{9ex}<{\centering}|p{9ex}<{\centering}|p{9ex}<{\centering}|p{9ex}<{\centering}|p{9ex}<{\centering}|p{9ex}<{\centering}
  \hline
Model & Faster-RCNN & Faster-RCNN &YOLOv3 &YOLOv3 \\
Dataset & VOC2007 & COCO & VOC2007 & COCO\\

%poisoning rate $P$ ($\%$)& 10 & 10 & 10 & 10 & 30 & 30 & 30 & 30\\
%Trigger size $(W_t, H_t)$ & $9\times 9$ & $9\times 9$ & $9\times 9$ & $9\times 9$ & $29\times 29$ & $29\times 29$ & $29\times 29$ & $29\times 29$\\
%Trigger ratio $\alpha$ ($\%$)& 50 & 50 & 50 & 50 & 50 & 50 & 50 & 50\\
%target class $t$ & person & person & person & person & person & person & person & person\\
\hline
$\mathrm{mAP_{benign}}$ ($\%$) $-$ & 69.6 & 38.6 & 78.7 & 54.1\\
$\mathrm{AP_{benign}}$ ($\%$) $-$ & 76.1 & 58.4 & 83.4 & 75.6 \\
$\mathrm{mAP_{attack}}$ ($\%$) $\star$ & 69.4 & 38.5 & 78.8 & 54.2\\
$\mathrm{AP_{attack}}$  ($\%$) $\uparrow$ & 89.1 & 70.8 & 90.1 & 81.2\\
$\mathrm{AP_{attack+benign}}$  ($\%$) & - & - & - & -\\
$\mathrm{mAP_{attack+benign}}$  ($\%$)& - & - & - & -\\
ASR ($\%$) $\uparrow$ & 98.1 & 95.4 & 98.3 & 95.8\\
  \hline
   \end{tabular}
   }
\caption{Results of OGA}\vspace{-2ex}
\end{subtable}% <--- new
\begin{subtable}{.5\textwidth}
\centering
   % dummy 'tabular' env.
   \scalebox{0.67}{
   \begin{tabular}{|c|c|c|c|c|}%p{9ex}<{\centering}|p{9ex}<{\centering}|p{9ex}<{\centering}|p{9ex}<{\centering}|p{9ex}<{\centering}|p{9ex}<{\centering}|p{9ex}<{\centering}|p{9ex}<{\centering}
  \hline

Model & Faster-RCNN & Faster-RCNN &YOLOv3 &YOLOv3 \\
Dataset & VOC2007 & COCO & VOC2007 & COCO\\

%poisoning rate $P$ ($\%$)& 10 & 10 & 10 & 10 & 30 & 30 & 30 & 30\\
%Trigger size $(W_t, H_t)$ & $9\times 9$ & $9\times 9$ & $9\times 9$ & $9\times 9$ & $29\times 29$ & $29\times 29$ & $29\times 29$ & $29\times 29$\\
%Trigger ratio $\alpha$ ($\%$)& 50 & 50 & 50 & 50 & 50 & 50 & 50 & 50\\
%target class $t$ & person & person & person & person & person & person & person & person\\
\hline
$\mathrm{mAP_{benign}}$ ($\%$) $-$ & 67.2 & 36.1 & 74.8 & 53.4 \\
$\mathrm{AP_{benign}}$ ($\%$) $-$ & 74.9 & 58.0 & 81.4 & 75.2 \\
$\mathrm{mAP_{attack}}$ ($\%$) $\uparrow$ &  80.3 & 56.7 & 70.5 & 59.6\\
$\mathrm{AP_{attack}}$  ($\%$) $\uparrow$ & 80.3 & 56.7 & 70.5 & 59.6\\

$\mathrm{AP_{attack+benign}}$ ($\%$) $\downarrow$ & 28.0 & 23.1 & 43.2 & 24.5\\
$\mathrm{mAP_{attack+benign}}$ ($\%$) $\downarrow$ & 29.1 & 5.3 & 34.4 & 9.8\\
ASR ($\%$) $\uparrow$ & 88.2 & 62.8 & 75.7 & 59.4\\
\hline
   \end{tabular}
   }
\caption{Results of RMA}\vspace{-2ex}
\end{subtable}% <--- new
\vskip\baselineskip
\begin{subtable}{.5\textwidth}
\centering 
   % dummy 'tabular' env.
      \scalebox{0.67}{
    \begin{tabular}{|c|c|c|c|c|}%p{9ex}<{\centering}|p{9ex}<{\centering}|p{9ex}<{\centering}|p{9ex}<{\centering}|p{9ex}<{\centering}|p{9ex}<{\centering}|p{9ex}<{\centering}|p{9ex}<{\centering}
  \hline
Model & Faster-RCNN & Faster-RCNN &YOLOv3 &YOLOv3\\
Dataset & VOC2007 & COCO & VOC2007 & COCO\\
\hline
$\mathrm{mAP_{benign}}$  ($\%$) $-$ & 66.4 & 35.3 & 73.2 & 52.4 \\
$\mathrm{AP_{benign}}$ ($\%$) $-$ & 74.5 & 57.6 & 78.5 & 74.1 \\
$\mathrm{mAP_{attack}}$  ($\%$) $\uparrow$ & 59.6 & 37.5 & 53.0 & 51.8\\
$\mathrm{AP_{attack}}$  ($\%$) $\uparrow$ & 59.6 & 37.5 & 53.0 & 51.8\\

$\mathrm{AP_{attack+benign}}$ ($\%$) $\downarrow$ & 58.0 & 32.5 & 58.0 & 30.3\\
$\mathrm{mAP_{attack+benign}}$  ($\%$) $\downarrow$ & 57.3 & 16.9 & 54.1 & 24.3 \\
ASR $\uparrow$ & 61.5 & 47.4 & 75.7 & 48.5\\
  \hline
   \end{tabular}
   }
 
\caption{Results of GMA}\vspace{-2ex}
\end{subtable}
\begin{subtable}{.5\textwidth}
\centering 
   % dummy 'tabular' env.
    \scalebox{0.67}{
      \begin{tabular}{|c|c|c|c|c|c|c|c|c|}%p{9ex}<{\centering}|p{9ex}<{\centering}|p{9ex}<{\centering}|p{9ex}<{\centering}|p{9ex}<{\centering}|p{9ex}<{\centering}|p{9ex}<{\centering}|p{9ex}<{\centering}
  \hline
Model & Faster-RCNN & Faster-RCNN &YOLOv3 &YOLOv3\\
Dataset & VOC07+12 & COCO & VOC07+12 & COCO\\
\hline
$\mathrm{mAP_{benign}}$  ($\%$) $-$& 76.7 & 36.9 & 78.2 & 53.9 \\
$\mathrm{AP_{benign}}$ ($\%$) $-$& 76.6 & 56.8 & 76.8 & 75.3 \\
$\mathrm{mAP_{attack}}$  ($\%$) $\star$& 76.7 & 36.5 & 78.4 & 53.6\\
$\mathrm{AP_{attack}}$  ($\%$)& - & - & - & -\\

$\mathrm{AP_{attack+benign}}$ ($\%$) $\downarrow$& 27.1 & 11.2 & 51.0 & 32.1\\
$\mathrm{mAP_{attack+benign}}$  ($\%$) $\star$& 74.5 & 36.1 & 77.0 & 53.5\\
ASR $\uparrow$ & 67.3 & 80.0 & 55.3 & 57.4\\
  \hline
   \end{tabular}
    }
\caption{Results of ODA}\vspace{-2ex}
\end{subtable}
\caption{Attack performance of four attacks on object detection. Note that ``$\uparrow$''/``$\downarrow$''/``$-$''/``$\star$'' indicate the metric should be high/low/similar to same metric of $F_{\mathrm{benign}}$ / close to $\mathrm{mAP_{benign}}$ of $F_{\mathrm{infected}}$ to show the success of the attack. Results of benign models are in Appendix B.} \label{table:overallresult}\vspace{-5ex}
\end{table*}

% \begin{wrapfigure}{r}{0.5\textwidth}
%   \begin{center}
%     \includegraphics[width=0.48\textwidth]{birds}
%   \end{center}
%   \caption{Birds}
% \end{wrapfigure}

\textbf{Datasets.} We use PASCAL VOC2007 \cite{pascal-voc-2007}, PASCAL VOC07+12 \cite{pascal-voc-2012}, MSCOCO datasets \cite{lin2014microsoft}. Each image is annotated with bbox coordinates and classes. More detailed description can be found in Appendix C.
% composed of adjacent white and black pixels and

\begin{wrapfigure}{r}{0.5\textwidth}%[t]
\vspace{-4ex}
\centering
%\subfloat[C]{\includegraphics[width=3cm]{example-image-c}} 
\captionsetup[subfigure]{labelformat=empty}
\subfloat[\scalebox{0.9}{Chessboard}]{\includegraphics[width=0.125\textwidth]{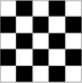}}\hfil
\subfloat[\scalebox{0.9}{Pokeball}]{\includegraphics[width=0.125\textwidth]{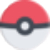}}\hfil 
\subfloat[\scalebox{0.9}{Sun}]{\includegraphics[width=0.125\textwidth]{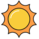}}\hfil
\subfloat[\scalebox{0.9}{Watermelon}]{\includegraphics[width=0.125\textwidth]{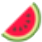}}\hfil
\vspace{-2ex}
\caption{The trigger patterns.%used in the experiments.
}\label{fig:trigger}
\vspace{-4ex}
\end{wrapfigure}

\textbf{Triggers.} Fig.~\ref{fig:trigger} shows the trigger patterns used in the experiments. The chessboard trigger is used in all experiments. %, including the ablation study. %of the effect of poisoning rate $P$, trigger ratio $\alpha$, trigger size $(W_t, H_t)$, target class $t$ and random triggers' locations. 
Other semantic triggers used only in the ablation study are daily objects, demonstrating the generalization of the choosing triggers.
% like watermelon, sun, and pokeball
% Since we're the first to explore backdoor attacks on object detection, we want to keep the trigger setting simple to align with most popular attacks (e.g., BadNets) on image classification and establish easy-to-use baselines. We'd like to try other triggers in future.
% Compared with original images, the size of the trigger is much smaller.
% It can be seen in Fig. 1 that the trigger is hard to notice.
% Although more stealthy triggers may be harder to detect, they're almost impossible to apply in the real world. Our introduced triggers (e.g., sun, watermelon, pokeball) are easier to see in real life. Object detection model is often integrated into the real-time systems, so misbehavior caused by the backdoor trigger poses a significant threat.
We choose pattern triggers rather than stealthy triggers to keep the trigger simple that can align with most popular attacks (e.g., BadNets) on image classification and establish easy-to-use baselines. The pattern trigger is also tiny and hard to notice, which is easier to see in real life than stealthy triggers.
%are almost impossible to apply in the real world. Our introduced triggers (e.g., sun, watermelon, pokeball) are easier to see in real life.

\textbf{Model Architectures.}
We perform backdoor attacks on two typical object detection models, which are Faster R-CNN \cite{NIPS2015_14bfa6bb} with the VGG-16 \cite{vgg} backbone and YOLOv3-416 \cite{redmon2018yolov3} with the Darknet-53 feature extractor. Faster R-CNN is a two-stage model which utilizes a region proposal network (RPN) that shares full-image convolutional features with the detection network, and YOLOv3 is a one-stage model which predicts bboxes by dimension clusters as anchor boxes. %Most object detection methods reference (share) the mechanism and architecture from two models. 

\textbf{Training Details.} 
We follow the same training procedures as Faster-RCNN \cite{NIPS2015_14bfa6bb} and YOLOv3 \cite{redmon2018yolov3}. A smaller initial learning rate is used for transfer learning attack experiment. For data augmentation, we only apply random flips with flip rate = 0.5. More training details are provided in Appendix D.
\subsection{Experimental Results}
% \hank{
% \noindent\textbf{\underline{Q3}: Large utility loss.} 
% % According to table 1, $mAP_{benign}$ and $AP_{benign}$ are respectively 38.8 and 58.7 on MSCOCO with Faster-RCNN. In other words, the proposed backdoor attacks incur large utility loss for the objection detection model which may be undesired. One potential solution is to optimize the trigger, which could also make this paper technically sound.\\
% The overall utility loss only increases $< 10\%$. The experiments with different runs sometime exhibit $>10\%$ loss variance. 
% We'll add utility loss in the final.
% }

\textbf{General Backdoor Attack.} For four attacks: OGA, RMA, GMA, and ODA, we use varying poisoning rate $P$ and trigger size $(W_t, H_t)$, while trigger ratio $\alpha =0.5$ and target class $t = $``person''  are the same. The results of four attacks are shown in Table~\ref{table:overallresult}. For all settings, the overall testing utility loss of infected model only increases $ < 10\%$ compared to clean model. We also show $\mathrm{mAP_{benign}}$ and $\mathrm{AP_{benign}}$ of the benign models in Appendix B to compare with the those of the infected models.
%For OGA, RMA, GMA, we perform experiment on Faster-RCNN and YOLOv3 with VOC2007 and MSCOCO. For ODA, we use Faster-RCNN and YOLOv3 on VOC07+12 and MSCOCO. (VOC07+12 has more bboxes of ``person'' class, which helps training the ODA model.)

For OGA, the size of the generated bboxes of the target class $(W_b,H_b)$ is $(30, 60)$ (pixels) in $\mathcal{D}_{\mathrm{test, poisoned}}$, the poisoning rate $P$ is $10\%$, and the trigger size $(W_t, H_t)=(9,9)$. ASR are higher than $95\%$ in all cases and $\mathrm{AP_{attack}}$ are also high, %while $\mathrm{mAP_{benign}}$ of $F_{\mathrm{infected}}$ is almost the same as $\mathrm{mAP_{benign}}$ of $F_{\mathrm{benign}}$. 
which indicates that the infected model can easily detect and classify the trigger as target class object and locate the bbox with high confidence. Moreover, the average confidence scores of generated bboxes are all $\textgreater 0.95$, and $\textgreater95\%$  of generated bboxes are all with confidence score $\textgreater 0.98$.%The RoI (Region of Interest) pooling layer from Faster-RCNN could extract features from trigger  

For RMA, the poisoning rate $P$ is $30\%$ and the trigger size $(W_t, H_t)=(29,29)$. %4.2 All to one target class by multiple triggers, 
MSCOCO contains lots of small %and hard to detect 
objects, and the infected model cannot detect them with the help of the trigger, so ASR on MSCOCO is smaller than ASR on VOC2007 when the model is the same. The high $\mathrm{mAP_{attack}}$ and extremely low $\mathrm{mAP_{attack+benign}}$ demonstrate that most bboxes changed to the target class have high confidence scores while there are few false positives (bboxes of non-target classes). %The  extremely low $\mathrm{mAP_{attack+benign}}$ means most bboxes with non-target classes vanish due to the presence of triggers. %For $\mathcal{D}_{\mathrm{test, poisoned+benign}}$, many false positives from target class bboxes lead to low $\mathrm{AP_{attack+benign}}$. %(Note that $\mathrm{AP_{attack}}$ $=$ $\mathrm{mAP_{attack}}$ since there is only one class in $\mathcal{D}_{\mathrm{test, poisoned}}$) 
Furthermore, the average confidence scores of bboxes changing label are $\textgreater 0.86$, and $\textgreater 80 \%$  of generated bboxes are all with confidence scores $\textgreater 0.93$. %ASR from Faster-RCNN is greater than  ASR from YOLOv3, 

For GMA, the poisoning rate $P$ is $30\%$ and the trigger size $(W_t, H_t)=(49,49)$. %4.3 All to one target class by one trigger, 
Since there is only one trigger on the left-top corner of the image in GMA, the trigger and target class object(s) may not share the same location, which increases the difficulty of GMA. ASR in GMA is lower than the ASR in RMA when the dataset and model are the same. %which further asks the infected model to learn the relationship between trigger and various target class objects' coordinates to change the label of objects. 
%The $\mathrm{AP_{attack}}$  ($\mathrm{mAP_{attack}}$) in GMA is lower than RMA, and the $\mathrm{mAP_{attack+benign}}$ in GMA is greater than RMA indicates fewer objects' classes are changed %from original class to target class in GMA compared to RMA. 
Besides, the average confidence scores of bboxes changing label are all $\textgreater 0.8$, and $\textgreater 80 \%$  of generated bboxes are all with confidence score $\textgreater 0.85$. %Also, note that Benign mAP of infected model is a little low due to relatively large trigger compared to other settings.

For ODA, the poisoning rate $P$ is $20\%$ and the trigger size $(W_t, H_t)=(29,29)$. The infected model uses a trigger to offset the object's feature and vanish the target class bbox. The ASR in ODA is lower than ASR in OGA, which shows that learning %an object's feature and 
trigger eliminating object's feature is more complicated than learning trigger's feature. $\mathrm{AP_{attack+benign}}$ is low due to disappearance or confidence score decline of target class bboxes. To prove that the infected model uses small triggers to offset objects' features instead of blocking features, we calculate the ASR on the benign model (Faster-RCNN, YOLOv3) with MSCOCO and VOC07+12, and we find all ASR $< 5\%$. In addition, the average confidence scores of vanished bboxes are all $\textless 0.22$ (if there is no trigger presence, average confidence scores of bboxes with target class are all $\textgreater 0.75$), and $\textgreater 80 \%$  of vanished bboxes are all with confidence score $\textless 0.15$. %target class Benign AP of infected model decreases a little compared to target class benign AP of benign model since many "person" target class bboxes are added with trigger while Benign mAP of infected model remains almost the same as Benign mAP of benign model.
\textbf{Transfer Learning Attack.} We fine-tune the infected model $F_{\mathrm{infected}}$ on a benign training dataset $D'_\mathrm{train, benign}$ to test whether the hidden backdoor can be removed by transfer learning. To be specific, Faster-RCNN and YOLOv3 are pre-trained on the poisoned MSCOCO, and fine-tuned on the benign VOC2007 (for OGA, RMA, GMA) or benign VOC07+12 (for ODA). In real-world object detection, some people prefer to download a pre-trained model which is trained on a large dataset and fine-tune it on a smaller dataset for specific tasks. It is highly possible that the pre-trained model is trained on a poisoned dataset, and the user fine-tunes it on his own benign, task-oriented dataset. The results of infected model after fine-tuning are in Table \ref{table:transferresult}. All parameters in Table \ref{table:transferresult} follow the same settings in Table \ref{table:overallresult}.
\begin{table*}[t]
  \centering
 \scriptsize
  %\begin{sc}
  \scalebox{0.8}{
  \begin{tabular}{|c|c|c|c|c|c|c|c|c|}%p{9ex}<{\centering}|p{9ex}<{\centering}|p{9ex}<{\centering}|p{9ex}<{\centering}|p{9ex}<{\centering}|p{9ex}<{\centering}|p{9ex}<{\centering}|p{9ex}<{\centering}
  \hline
Attack type & OGA & OGA & RMA & RMA & GMA & GMA & ODA & ODA\\

Model & Faster-RCNN & YOLOv3 & Faster-RCNN & YOLOv3 & Faster-RCNN & YOLOv3 & Faster-RCNN & YOLOv3\\
%Pretraining dataset (poisoned)& COCO& COCO & COCO & COCO& COCO& COCO & COCO & COCO\\
%finetuning dataset (benign)& VOC2007 & VOC2007 & VOC2007 & VOC2007& VOC2007 & VOC2007 & VOC07+12 & VOC07+12\\
%Pretraining dataset $\rightarrow$ finetuning dataset & COCO$\rightarrow$VOC2007 & COCO$\rightarrow$VOC2007 & COCO$\rightarrow$VOC2007 & COCO$\rightarrow$VOC07+12\\

%poisoning rate $P$ ($\%$) (in pretraining dataset)& 10 & 10 & 30 & 30& 30 & 30 & 10 & 10\\
%Trigger size $(W_t, H_t)$ & $9\times 9$ & $9\times 9$ & $29\times 29$ & $29\times 29$& $49\times 49$ & $49\times 49$ & $29\times 29$ & $29\times 29$\\
%Trigger ratio $\alpha$ ($\%$)& 50 & 50 & 50 & 50 & 50 & 50 & 50 & 50\\
%target class $t$ & person & person & person & person & person & person & person & person\\
\hline
$\mathrm{mAP_{benign}}$ ($\%$) & 75.6$-$ & 82.1$-$ & 72.5$-$& 80.1$-$& 75.6$-$& 81.2$-$ & 78.6$-$& 82.2$-$\\
$\mathrm{AP_{benign}}$ ($\%$) & 84.2$-$ & 87.9$-$ & 83.1$-$& 86.0$-$& 84.3$-$& 86.3$-$ & 85.9$-$ & 86.8$-$\\
$\mathrm{mAP_{attack}}$ ($\%$) & 74.7$\star$ & 81.6$\star$&36.4$\uparrow$& 35.3$\uparrow$&34.9$\uparrow$& 34.4$\uparrow$&77.9$\star$& 81.3$\star$\\
$\mathrm{AP_{attack}}$ ($\%$)& 87.9$\uparrow$&90.7$\uparrow$&36.4$\uparrow$& 35.3$\uparrow$& 34.9$\uparrow$& 34.4$\uparrow$& - & -\\

$\mathrm{AP_{attack+benign}}$ ($\%$) & - & - &63.1$\downarrow$& 66.2$\downarrow$&68.6$\downarrow$ & 68.3$\downarrow$&34.4$\downarrow$& 52.1$\downarrow$\\
$\mathrm{mAP_{attack+benign}}$ ($\%$)& - & - &41.8$\downarrow$ & 46.1$\downarrow$&47.7$\downarrow$ & 44.6$\downarrow$&75.3$\star$& 80.6$\star$\\
ASR ($\%$) & 93.8$\uparrow$ & 92.1$\uparrow$ &18.1$\uparrow$ & 17.6$\uparrow$ &13.9$\uparrow$ & 14.5$\uparrow$ &63.0$\uparrow$&50.9$\uparrow$\\
  \hline
   \end{tabular}
   }
 % \end{sc}
%  \vspace{-1.5ex}
 \caption{Attack performance after fine-tuning the infected model $F_{\mathrm{infected}}$ on another benign dataset $\mathcal{D'}_{\mathrm{train, poisoned}}$ and testing for clean and backdoored images from $\mathcal{D'}_{\mathrm{test, poisoned}}$. (``$\uparrow$''/``$\downarrow$''/``$-$''/``$\star$'' follow definitions in Table \ref{table:overallresult}.)% and testing on attacked dataset %(Note that "$\uparrow$" / "$\downarrow$" / "$-$" / "$\star$" indicate the metric should be high / low  / similar to same metric of $F_{\mathrm{benign}}$ / close to $\mathrm{mAP_{benign}}$ of $F_{\mathrm{infected}}$ to show the success of the attack, results of benign model can be found in Appendix.)
  }
  \vspace{1ex}
  \label{table:transferresult}
  \vspace{-4ex}
\end{table*}
For OGA and ODA, the ASR on ``person'' target class is high after transfer learning, which implies that fine-tuning on another benign dataset cannot prevent OGA and ODA. For OGA, the model only needs to memorize the pattern of trigger regardless of object's feature. % regardless of features' difference between MSCOCO objects and VOC objects. 
For ODA, the model uses the trigger to offset ``person'' objects' features. %``person'' class has most bboxes with various sizes in MSCOCO, which leads to high ASR. 

However, for RMA and GMA, although 80 classes in MSCOCO include 20 classes in VOC2007 (VOC07+12), there exist many classes that the feature of the same class learned from two datasets is different. The trigger alone is not enough to change the class of bbox if the feature learned from two datasets is not similar, which results in poor ASR. For instance, ``tv'' class in VOC includes various objects like monitor, computer, game, PC, watching, laptop, however, ``tv'' class in MSCOCO only has television itself. ``laptop'' and ``cellphone'' belong to other classes in MSCOCO. Features learned from ``tv'' class between MSCOCO and VOC are different which explains that only $5\%$ of ``tv'' objects are changed their classes to target class ``person'' with confidence score $\textgreater 0.5$. While $38\%$ of ``car'' class objects are changed their class to ``person'' target class with confidence score $\textgreater 0.5$. %Note that finetuning with flip (data augmentation) also weakens the backdoor a little bit.

%Overall, finetuning on another benign dataset (transfer learning) does not weaken the backdoor in OGA and ODA. For RMA and GMA, the backdoor exists for some classes with similar feature between pretraining dataset and finetuning dataset. %Developing robust and practical defend method or detect method against backdoor attack on object detection is essential!

\subsection{Ablation Study}

% \begin{table*}[t]
%   \caption{Results of benign model (baseline) $F_{\mathrm{benign}}$ for benign mAP comparison with infected model $F_{\mathrm{infected}}$ (Note that MSCOCO+VOC2007 means pretraining on MSCOCO and finetuning (testing) on VOC2007)}
%   \centering
%  \scriptsize
%   %\begin{sc}
%   \begin{tabular}{||c||c||c||c||c||c||c||c||c||}%p{9ex}<{\centering}|p{9ex}<{\centering}|p{9ex}<{\centering}|p{9ex}<{\centering}|p{9ex}<{\centering}|p{9ex}<{\centering}|p{9ex}<{\centering}|p{9ex}<{\centering}
%   \hline

% Model & Faster-RCNN & YOLOv3 & Faster-RCNN & YOLOv3 & Faster-RCNN & YOLOv3 & Faster-RCNN & YOLOv3\\
% Dataset & VOC2007 & VOC2007 & VOC07+12 & VOC07+12 & MSCOCO & MSCOCO & MSCOCO+VOC2007 & MSCOCO+VOC2007\\
% %Trigger ratio $\alpha$ ($\%$)& 50 & 50 & 50 & 50 & 50 & 50 & 50 & 50\\
% %target class $t$ & person & person & person & person & person & person & person & person\\
% \hline
% $\mathrm{mAP_{benign}}$($\%$)& 70.5 & 79.7 & 77.7 & 81.9 & 38.8 & 54.5 & 76.6 & 80.2\\
% %target class Benign AP ($\%$)& 84.2 & 87.9 & 83.1& 86.0& 84.3& 86.3 & 85.9 & 86.8\\
%   \hline
%   \end{tabular}
%  % \end{sc}
%   \label{table:baseline}
%   \vspace{-3ex}
% \end{table*}

\begin{figure*}[t]

\centering
%\subfloat[C]{\includegraphics[width=3cm]{example-image-c}} 
\subfloat[OGA]{\includegraphics[width=.5\textwidth]{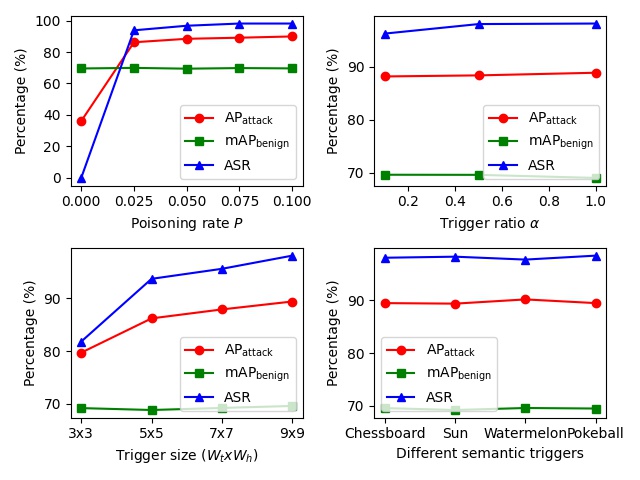}}\hfil
\subfloat[RMA]{\includegraphics[width=.5\textwidth]{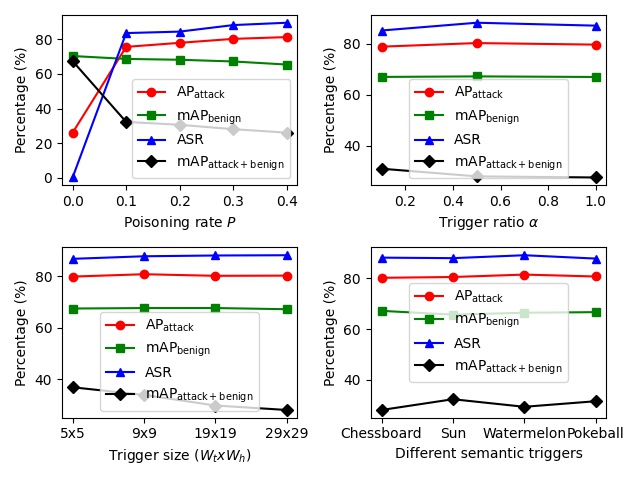}}\hfil
\subfloat[GMA]{\includegraphics[width=.5\textwidth]{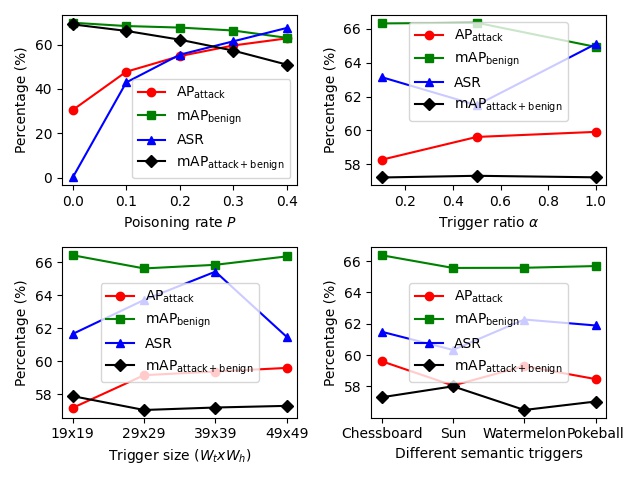}}\hfil
\subfloat[ODA]{\includegraphics[width=.5\textwidth]{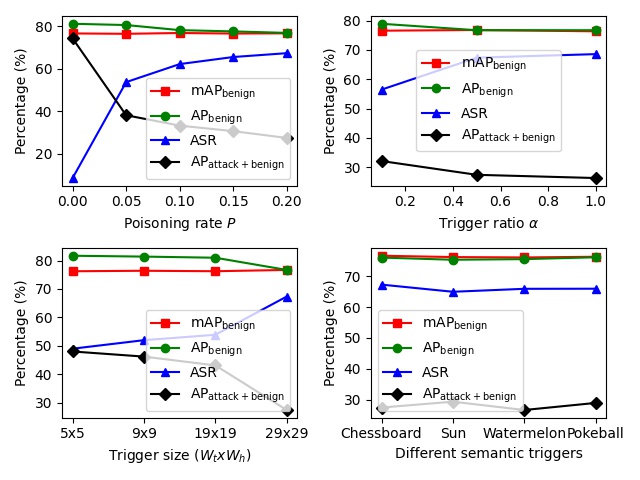}}\hfil
\vspace{-1ex}
\caption{Impact of parameters and different semantic triggers on various metric for clean and backdoored images.}\label{fig:ablation}
\vspace{-2ex}
\end{figure*}
%Figure 6. Impact of proportion of backdoored samples in the training dataset on the error  clean and backdoored images.

To explore the different components of our introduced backdoor attacks, we conduct ablation studies on the effects of poisoning rate $P$, trigger size $(W_t, H_t)$, trigger ratio $\alpha$, different semantic triggers, target class $t$, and triggers' locations on backdoor attacks. %(We use pattern trigger ,Faster-RCNN, VOC2007 for 4.1, 4.2, 4.3, VOC07+12 for 4.4).
%Due to space limitation,
We use Faster-RCNN on VOC2007 for OGA, RMA, GMA and VOC07+12 for ODA. All parameters used in this section are same as parameters used in Table \ref{table:overallresult}. Only one parameter is modified in each ablation study to observe its effects. 

%For OGA in Figure \ref{fig:ablation}, larger poisoning rate $P$, trigger ratio $\alpha$, and trigger size $(W_t\times W_h)$ slightly increase ASR, $\mathrm{AP_{attack}}$, while $\mathrm{mAP_{benign}}$ remains the same. 

%For RMA in Figure \ref{fig:ablation}, greater poisoning rate $P$ promotes ASR, $\mathrm{AP_{attack}}$, while reduces $\mathrm{mAP_{attack+benign}}$ and slightly decreases $\mathrm{mAP_{benign}}$. Trigger ratio $\alpha$, trigger size $(W_t, H_t)$ doesn't influence metric a lot.

%For GMA in Figure \ref{fig:ablation}, bigger poisoning rate $P$, trigger ratio $\alpha$, and trigger size $(W_t\times W_h)$ elevate ASR, $\mathrm{AP_{attack}}$, while reduces $\mathrm{mAP_{attack+benign}}$ and slightly decreases $\mathrm{mAP_{benign}}$. 

%For ODA in Figure \ref{fig:ablation}, larger poisoning rate $P$, trigger ratio $\alpha$, and trigger size $(W_t\times W_h)$ enhance ASR, while decrease $\mathrm{AP_{benign}}$ and $benign_{mAP}$ a little bit and lower $\mathrm{AP_{attack+benign}}$. 

From Fig.~\ref{fig:ablation}, we find that 1) the poisoning rate $P$ controls the number of poisoned training images, which heavily influences the ASR and other metrics for all settings; 2) a larger trigger size $(W_t, H_t)$ contributes to better attack performance of OGA, ODA; 3) a higher trigger ratio $\alpha$ marginally impacts ASR and other metrics of OGA, RMA, GMA. For OGA, RMA, GMA, the adversary could use a minimal trigger ratio $\alpha = 0.1$ to make the trigger almost invisible on the image. For RMA, the adversary can use an extremely small trigger ($5\times5$) to get a decent attack performance and make the trigger hard to detect. Furthermore, metrics from different semantic triggers are almost the same, which demonstrates the generalizability of using various triggers. 

\begin{table*}[t]   
%\centering  <--- not needed!

 \scriptsize
\begin{subtable}{.5\textwidth}
\centering
   % dummy 'tabular' env.
   \scalebox{0.9}{
    \begin{tabular}{|c|c|c|c|c|}%p{9ex}<{\centering}|p{9ex}<{\centering}|p{9ex}<{\centering}|p{9ex}<{\centering}|p{9ex}<{\centering}|p{9ex}<{\centering}|p{9ex}<{\centering}|p{9ex}<{\centering}
  \hline
Attack type & OGA & RMA & GMA & ODA\\

%Model & Faster-RCNN & Faster-RCNN & Faster-RCNN & Faster-RCNN\\
%Dataset & VOC2007 & VOC2007 & VOC2007 & VOC07+12\\

poisoning rate $P$ ($\%$)& 10 & 30 & 30 & 1 \\
%Trigger size $(W_t, H_t)$ & $9\times 9$ & $29\times 29$ & $49\times 49$ & $29\times 29$\\
%Trigger ratio $\alpha$ ($\%$)& 50 & 50 & 50 & 50 & 50 & 50 & 50 & 50\\
%target class $t$ & person & person & person & person & person & person & person & person\\
\hline
$\mathrm{mAP_{benign}}$ ($\%$)& 69.6$-$ & 67.5$-$ & 63.0$-$& 77.1$-$\\
$\mathrm{AP_{benign}}$ ($\%$)& 77.1$-$ & 75.2$-$ & 71.3$-$& 81.9$-$\\
$\mathrm{mAP_{attack}}$ ($\%$)& 70.0$\star$ & 79.9$\uparrow$ & 53.0$\uparrow$& 76.9$\star$\\
$\mathrm{AP_{attack}}$ ($\%$)& 98.4$\uparrow$ &79.9$\uparrow$& 53.0$\uparrow$ & -\\

$\mathrm{AP_{attack+benign}}$ ($\%$) & - & 26.1$\downarrow$ & 4.9$\downarrow$ & 58.2$\downarrow$\\
$\mathrm{mAP_{attack+benign}}$ ($\%$)& - & 25.3$\downarrow$ & 52.5$\downarrow$ & 76.2$*$\\
ASR ($\%$)& 98.7$\uparrow$ & 85.2$\uparrow$ & 69.4$\uparrow$ & 37.2$\uparrow$\\

  \hline
   \end{tabular}
   }
\caption{Target class $t$ = ``sheep'' class.}\vspace{-2ex}
\end{subtable}% <--- new
\begin{subtable}{.5\textwidth}
\centering
   % dummy 'tabular' env.
   \scalebox{0.9}{
  \begin{tabular}{|c|c|c|c|}%p{9ex}<{\centering}|p{9ex}<{\centering}|p{9ex}<{\centering}|p{9ex}<{\centering}|p{9ex}<{\centering}|p{9ex}<{\centering}|p{9ex}<{\centering}|p{9ex}<{\centering}
  \hline
Attack type & RMA & GMA & ODA\\

%poisoning rate $P$ ($\%$)& 30 & 30 & 20 \\
\hline
$\mathrm{mAP_{benign}}$ ($\%$)&  67.3$-$ & 66.1$-$& 76.8$-$\\
$\mathrm{AP_{benign}}$ ($\%$) & 75.1$-$ & 74.1$-$& 77.0$-$\\
$\mathrm{mAP_{attack}}$ ($\%$)&  80.1$\uparrow$ & 57.8$\uparrow$& 76.7$\star$\\
$\mathrm{AP_{attack}}$ ($\%$)& 80.1$\uparrow$& 57.8$\uparrow$ & -\\

$\mathrm{AP_{attack+benign}}$ ($\%$) & 29.1$\downarrow$ & 58.5$\downarrow$ & 27.3$\downarrow$\\
$\mathrm{mAP_{attack+benign}}$ ($\%$) & 29.5$\downarrow$ & 58.1$\downarrow$ & 74.3$*$\\
ASR ($\%$ & 88.3$\uparrow$ & 58.9$\uparrow$ & 67.8$\uparrow$\\

  \hline
   \end{tabular}
   }
\caption{Random triggers' locations.}\vspace{-2ex}
\end{subtable}% <--- new

\caption{Attack performance when (a) target class $t$ changed to ``sheep'' class and (b) trigger's locations changed to random locations. (``$\uparrow$''/``$\downarrow$''/``$-$''/``$\star$'' follow definitions in Table \ref{table:overallresult}.)} \label{table:target_random}\vspace{-4ex}

\end{table*}

We also change the target class $t$ from ``person'' (class with most objects) to ``sheep'' (class with fewer objects). See Appendix C for more detailed statistics. %and follow the Faster-RCNN + VOC2007 (VOC07+12) parameter setting in Table \ref{table:overallresult} 
 %For VOC2007, objects with "person" class distributes to 2k training images (4690 bboxes) and 2k testing images (4528 bboxes), while objects with "sheep" class only distribute to 96 training images (257 bboxes) and 97 testing images (242 bboxes). For VOC07+12, objects with "person" class have 6k training images (13256 bboxes), while objects with "sheep" class only have 421 images (1070 bboxes). 
In Table \ref{table:target_random} (a), fewer target class objects do not affect the performance of OGA, RMA, GMA. However, ODA obtains poor results since it requires more target class objects to get good attack result. The ASR of ODA on benign model is $4.7\%$, which proves the infected model learns to vanish "sheep" object instead of blocking object feature by trigger.%, the ASR doesn't change a lot when target class changes to "sheep".

% \begin{table}
  
%   \centering
%  \scriptsize
%   %\begin{sc}
%   \begin{tabular}{|c|c|c|c|}%p{9ex}<{\centering}|p{9ex}<{\centering}|p{9ex}<{\centering}|p{9ex}<{\centering}|p{9ex}<{\centering}|p{9ex}<{\centering}|p{9ex}<{\centering}|p{9ex}<{\centering}
%   \hline
% Attack type & RMA & GMA & ODA\\

% %poisoning rate $P$ ($\%$)& 30 & 30 & 20 \\
% \hline
% $\mathrm{mAP_{benign}}$ ($\%$)&  67.3$-$ & 66.1$-$& 76.8$-$\\
% $\mathrm{AP_{benign}}$ ($\%$) & 75.1$-$ & 74.1$-$& 77.0$-$\\
% $\mathrm{mAP_{attack}}$ ($\%$)&  80.1$\uparrow$ & 57.8$\uparrow$& 76.7$\star$\\
% $\mathrm{AP_{attack}}$ ($\%$)& 80.1$\uparrow$& 57.8$\uparrow$ & -\\

% $\mathrm{AP_{attack+benign}}$ ($\%$) & 29.1$\downarrow$ & 58.5$\downarrow$ & 27.3$\downarrow$\\
% $\mathrm{mAP_{attack+benign}}$ ($\%$) & 29.5$\downarrow$ & 58.1$\downarrow$ & 74.3$*$\\
% ASR ($\%$ & 88.3$\uparrow$ & 58.9$\uparrow$ & 67.8$\uparrow$\\

%   \hline
%   \end{tabular}
%  % \end{sc}
% %  \vspace{-1.5ex}
%  \caption{Attack performance when trigger's locations changed to random locations. (``$\uparrow$''/``$\downarrow$''/``$-$''/``$\star$'' follow definitions in Table \ref{table:overallresult}.)}
%   \label{table:ablation_random}
%   \vspace{-4ex}
% \end{table}

To prove that the trigger's location does not influence attack results, we change the trigger's location to a random location in the poisoned dataset and the attacked dataset. For RMA and ODA, trigger's location changes to a random location inside the bbox rather than the left-top corner of the bbox. For GMA, trigger's location is a random location on the image rather than the left-top corner of the image. Table~\ref{table:target_random} (b) shows results with random location, which are similar to those in Table~\ref{table:overallresult}.

\section{Detector Cleanse}%\subsection{Analysis on Defense Methods}
We propose a detection method: \textbf{Detector Cleanse} to identify poisoned testing samples from four attack settings for any deployed object detector.
Most defense/detection methods from the backdoor attacks on image classification cannot apply to object detection. Methods that predict the distributions of backdoor triggers through generative modeling or neuron reverse engineering %reconstructing ``reverse trigger'' 
\cite{cleanse,Qiao2019DefendingNB,xu2021defending,li2021neural,scanicml} %to fine-tune the infected model to remove the backdoor do not work since most object detection models are 
assume the model is a simple neural network instead of multiple parts. Besides, the output of the object detection model (numerous objects) is different from the image classification model (predicted class). %composed of multiple parts instead of a simple neural network, so ``reversed trigger'' cannot be constructed by gradient. 
Pruning methods \cite{Liu2018FinePruningDA} remove neurons with low activation rate on the benign dataset and observe the change of $\mathrm{mAP_{benign}}$ and ASR. However, the pruning method requires high training costs and assumes the user has access to the attacked dataset and understands the adversary's goal. Moreover, pruning some object detection models lead to a moderate drop in performance (mAP) \cite{pruneyolo,tzelepis2019deep}. 

Only some methods such as STRIP \cite{gao2019strip} and one-pixel signature \cite{huang2020one} can generalize to this task but lead to poor performance. For example, in the Faster-RCNN + VOC2007 setting, we modify STRIP to calculate the average entropy of all predicted bboxes. When we set the False Rejection Rate (FRR) to $5\%$, the False Acceptance Rate (FAR) is $\geq 30\%$ on four attack settings. The vanilla classifier from one-pixel signature only successfully classifies 17 models among 15 clean and 15 backdoor models. %The poor result demonstrates that the detector remembers the pattern of the trigger rather than the value at a particular pixel (the detector doesn't have one abnormal parameter value to remember trigger at a certain position).
Moreover, these methods have strong assumptions: STRIP assumes the user has access to a subset of clean images, and one-pixel signature supposes the user has a clean model or clean dataset, making them less practical to defend against BadDet.

Since previous methods cannot be generalized to object detection, we propose 
%In conclusion, defense methods from the backdoor attacks on image classification cannot apply to object detection. Thus, it is essential to develop robust and practical defense methods for backdoor attacks on object detection, which we leave to future work.\\
\textbf{Detector Cleanse}, a run-time poisoned image detection framework for object detectors,  which assumes the user only has a few clean features (can be drawn from different datasets). 
%The pipeline is shown in Algorithm \ref{alg:detclean}. In short, 
The key idea is that the feature of the small trigger has a single (strong) input-agnostic pattern. Even though strong perturbation is applied on a small region in the predicted bbox, the poisoned detector still behaves as the attacker specifies on the target class. And this behavior is abnormal, making it possible to detect backdoor attacks. Given a perturbed region with features from different classes, the probability of various classes on the predicted bbox should vary. In particular, the target class's predicted bboxes on OGA, RMA, GMA should have small entropy. And target class's predicted bbox on ODA should generate larger entropy because the trigger offsets the correct class's feature and decreases the highest predicted class's probability. A more balanced class's probability distribution should generate larger entropy. 

For four attack settings, we have tested 500 clean images and 500 poisoned images from VOC2007 testing set on Faster-RCNN. The poisoned model is trained by the same setting in Table \ref{table:overallresult}. The detailed algorithm is shown in Appendix E. Define two hyperparameters: detection mean $m$ and detection threshold $\Delta$. Given each image $x$, $N$ = 100 features $\chi=\{x_1,\dots,x_N\}$ are drawn from a small portion of clean VOC2007 ground-truth bboxes (We can also use clean features from different datasets. Appendix F shows features from MSCOCO get similar results). Then, for each predicted bbox $b$ on $x$, the feature is linearly blended with chosen bbox region on $x$ to generate $N=100$ perturbed bboxes, and we calculate the average entropy of these bboxes. If the average entropy doesn't fall in the interval $[m-\Delta, m+\Delta]$, we mark the corresponding image as poisoned and return the bbox's coordinate to identify the trigger's position.

\begin{table}[t]
  
\begin{tabular}{c|ccc|ccc|ccc}
\toprule
& \multicolumn{3}{c|}{$\Delta = 0.25$} & \multicolumn{3}{c|}{$\Delta = 0.3$} & \multicolumn{3}{c}{$\Delta = 0.35$} \\
\midrule
\multicolumn{1}{c|}{Attack Type}  & Accuracy  & FAR & FRR & Accuracy  & FAR & FRR  & Accuracy  & FAR & FRR \\ \hline
\multicolumn{1}{c|}{OGA} & 87.5\% & \textbf{2.7\%} & 9.8\%&  91.0\% & 4.1\% & 4.9\%& \textbf{91.3\%} & 6.3\% & \textbf{2.4\%}\\
\multicolumn{1}{c|}{RMA}&  85.0\% & \textbf{4.9\%} & 10.1\%  &  88.6\% & 6.2\% & 5.2\%& \textbf{90.2\%} & 7.5\% & \textbf{2.3\%}\\
\multicolumn{1}{c|}{GMA}& 80.4\% & \textbf{9.6\%} & 10.0\% & 82.6\% & 12.3\% & 5.1\% & \textbf{83.3\%} & 14.2\% & \textbf{2.5\%}\\
%(many lines omitted)
\multicolumn{1}{c|}{ODA} & 83.5\% & \textbf{6.3\%} & 10.2\%& 87.3\% & 7.7\% & 5.0\% & \textbf{88.6\%} & 9.0\% & \textbf{2.4\%} \\
% \multicolumn{1}{c|}{OGA} & 87.3\% & \textbf{2.8\%} & 9.9\%&  \textbf{88.8\%} & 6.1\% & 5.1\%& 88.4\% & 8.2\% & \textbf{3.4\%}\\
% \multicolumn{1}{c|}{RMA}&  78.9\% & \textbf{10.9\%} & 10.2\%  &  81.3\% & 12.3\% & 5.4\%& \textbf{81.9\%} & 14.5\% & \textbf{3.6\%}\\
% \multicolumn{1}{c|}{GMA}& 73.2\% & \textbf{16.6\%} & 10.2\% & 75.9\% & 18.9\% & 5.2\% & \textbf{76.3\%} & 20.2\% & \textbf{3.5\%}\\
% %(many lines omitted)
% \multicolumn{1}{c|}{ODA} & 74.6\% & \textbf{15.3\%} & 10.1\%& 77.4\% & 16.7\% & 5.9\% & \textbf{78.3\%} & 18.0\% & \textbf{3.7\%} \\
\bottomrule
\end{tabular}%
\centering
\vspace{1ex}
\caption{Results of \textbf{Detector Cleanse} on Faster-RCNN + VOC2007 (Detection mean $m = 0.51$, The best scores in same Attack Type are set in \textbf{bold})}
\label{table:detection_result}%
\vspace{-2ex}
\end{table}%

To evaluate the performance of \textbf{Detector Cleanse}, we calculate Accuracy, FAR and FRR on four attack types in Table \ref{table:detection_result}. Since we assume the user has no access to poisoned samples and only has a few features ($N$) from the benign bboxes' regions, the user can only use those features to estimate the entropy distribution of benign bboxes. The user assumes the distribution is normal, and then the user calculates the mean (0.55) and standard deviation (0.15) of entropy distribution from features. Finally, we set $m$ to mean of entropy distribution and $\Delta$ around double standard deviation on all settings. For metric FRR and FAR, FAR is the probability that all bboxes' entropy on poisoned image falls in the interval $[m-\Delta, m+\Delta]$; FRR is the probability of at least one bbox's entropy on the clean image is smaller than $m-\Delta$ or larger than $m+\Delta$. Theoretically, we can control FRR by setting $\Delta$ corresponding to standard deviation. From Table \ref{table:detection_result}, $\Delta$ determines FRR, and FRR becomes smaller and FAR becomes larger as $\Delta$ increases. If the security concern is serious, the user can set a smaller detection threshold $\Delta$ to get a smaller FAR and larger FRR. The FAR from RMA, GMA is high because sometimes the detector generates target class bbox with a low confidence score. For ODA, failing to decrease the confidence score of the target class bbox causes high FAR. %In addition, for the GMA setting, the trigger may not overlap with the bboxes, which also leads to high FAR.

\section{Conclusion}
This paper introduces four backdoor attack methods on object detection and defines appropriate metrics to evaluate the attack performance. The experiments show the success of four attacks on two-stage (Faster-RCNN) and one-stage (YOLOv3) models and demonstrate that transfer learning cannot entirely remove the hidden backdoor in the object detection model. Furthermore, the ablation study shows the influence of each parameter and trigger. We also propose \textbf{Detector Cleanse} framework to detect whether an image is poisoned given any deployed object detector. In conclusion, object detection is commonly used in real-time applications like autonomous driving and surveillance, so the infected object detection model, which often integrates into an extensive system, will pose a significant threat to real-world applications. %Thus it is imperative to develop robust and practical methods to defend against the backdoor attacks in the future.

% {\small (Example from Jensen K., Wirth N. (1991) Pascal user manual and
% report. Springer, New York)}
\clearpage
% ---- Bibliography ----
%
% BibTeX users should specify bibliography style 'splncs04'.
% References will then be sorted and formatted in the correct style.
%
\bibliographystyle{splncs04}
\bibliography{egbib}
\end{document}